\documentclass[sigconf]{acmart}

\renewcommand\footnotetextcopyrightpermission[1]{}
\usepackage{algorithm}
\usepackage{algpseudocode}
\usepackage{amsmath}
\usepackage{amssymb}  
\usepackage{multirow} 

\usepackage{graphicx}
\usepackage{booktabs}
\usepackage{caption}
\usepackage{subcaption}
\usepackage{hyperref}
\usepackage{float}

\acmConference[Preprint]{arXiv}{arXiv}{2025} 
\acmYear{}
\copyrightyear{}

\begin{document}

\title[MOTGNN]{MOTGNN: Interpretable Graph Neural Networks for Multi-Omics Disease Classification}

\author{Tiantian Yang}
\authornote{Corresponding author}
\affiliation{
  \department{Mathematics and Statistical Science}
  \institution{University of Idaho}
  \city{Moscow}
  \state{Idaho}
  \country{USA}
}
\email{tyang@uidaho.edu}

\author{Zhiqian Chen}
\affiliation{
  \department{Computer Science and Engineering}
  \institution{Mississippi State University}
  \city{Starkville}
  \state{Mississippi} 
  \country{USA}
}      

\renewcommand{\shortauthors}{Yang and Chen}

\begin{abstract}
Integrating multi-omics data, such as DNA methylation, mRNA expression, and microRNA (miRNA) expression, offers a comprehensive view of the biological mechanisms underlying disease. However, the high dimensionality of multi-omics data, the heterogeneity across modalities, and the lack of reliable biological interaction networks make meaningful integration challenging. In addition, many existing models rely on handcrafted similarity graphs, are vulnerable to class imbalance, and often lack built-in interpretability, limiting their usefulness in biomedical applications. We propose Multi-Omics integration with Tree-generated Graph Neural Network (\textbf{MOTGNN}), a novel and interpretable framework for binary disease classification. MOTGNN employs eXtreme Gradient Boosting (XGBoost) for omics-specific supervised graph construction, followed by modality-specific Graph Neural Networks (GNNs) for hierarchical representation learning, and a deep feedforward network for cross-omics integration. Across three real-world disease datasets, MOTGNN outperforms state-of-the-art baselines by 5-10\% in accuracy, ROC-AUC, and F1-score, and remains robust to severe class imbalance. The model maintains computational efficiency through the use of sparse graphs and provides built-in interpretability, revealing both top-ranked biomarkers and the relative contributions of each omics modality. These results highlight the potential of MOTGNN to improve both predictive accuracy and interpretability in multi-omics disease modeling.
\end{abstract}


\maketitle

\section{Introduction}
A comprehensive understanding of complex diseases, such as cancer, cardiovascular disease, and neurodegenerative disorders, is critical for advancing public health and enabling precision medicine. These diseases are not driven by isolated molecular events but instead emerge from intricate interactions among genomic, epigenomic, and transcriptomic factors \citep{subramanian2020multi}. Recent advances in high-throughput biotechnologies have enabled the simultaneous collection of multiple molecular layers, such as gene expression (mRNA), epigenetic modifications (DNA methylation), and non-coding RNA activity (e.g., microRNA), from the same set of biological samples \citep{hood2013human}. 
These omics datasets capture intricate relationships among biological entities and molecular processes, offering insights into disease mechanisms. While individual omics types provide valuable information, each reflects distinct biological processes and contributes complementary information. Integrating multiple omics (multi-omics) data is therefore essential for comprehensively understanding the multi-layered systems underlying disease processes and progression \citep{waqas2024multimodal}. 
However, integrating heterogeneous omics data presents unique challenges. A fundamental issue is that the number of samples ($n$) is typically much smaller than the number of features ($p$). For example, a study may involve only a few hundred participants while measuring tens of thousands of molecular variables. This high dimensionality complicates model training and increases the risk of overfitting. Moreover, not all measured features are biologically relevant; noisy or redundant features can obscure meaningful patterns. Although biological interaction networks such as gene-gene or protein-protein interactions have been identified experimentally, current knowledge is constrained by limitations in experimental design and equipment. Consequently, there is a growing need for computational models that can reliably infer biologically meaningful structures and interactions directly from multi-omics data \citep{jiang2025network}.

Traditional machine learning and deep learning models have been widely applied to address the high dimensionality and complexity of omics data \citep{zaghlool2022review, ayman2023review}, but most assume a Euclidean data structure. Classical methods such as random forests \citep{breiman2001random} and gradient boosting algorithms like XGBoost \citep{chen2016xgboost} construct ensembles of decision trees and perform well for classification, yet they typically treat features as independent and fail to capture structural relationships among them. 
Deep learning models \citep{Lecun2015}, including deep feedforward networks (DFNs) \citep{goodfellow2016deep}, convolutional neural networks, and recurrent neural networks, can model nonlinear feature dependencies but are not designed to process non-Euclidean data structures such as graphs. These approaches typically require large training sets to prevent overfitting and are often criticized for their black-box nature, which limits interpretability, an essential requirement in biomedical research where transparency and reproducibility are critical. Furthermore, simply concatenating features from different omics modalities into a single input can obscure modality-specific relationships and introduce additional noise. 
Graph-based approaches provide a natural alternative for modeling complex biological systems. In omics applications, nodes can represent features (e.g., genes) or samples, and edges encode various relationships such as similarity, co-expression, or learned interactions. Graph Neural Networks (GNNs) \citep{bronstein2017geometric, wu2020comprehensive, zhang2020deep, zhou2020graph} have emerged as powerful tools for learning from graph-structured data by recursively aggregating information from each node's neighborhood. GNNs capture both local and global graph topological patterns and have shown increasing success in biomedical domains, including molecular modeling, drug discovery, and disease prediction \citep{zhang2021graph, johnson2024graph, waqas2024multimodal, zhong2025survey}. Foundational architectures such as message passing neural networks (MPNNs) \citep{gilmer2017neural}, graph convolutional networks (GCNs) \citep{kipf2017semi}, graph attention networks (GATs) \citep{petar2018graph}, GraphSAGE \citep{hamilton2017inductive}, and graph isomorphism networks (GINs) \citep{xu2019powerful} have advanced the expressiveness and scalability of GNN-based learning.

Recent GNN-based methods have aimed to integrate multi-omics data for disease classification, cancer subtype analysis, and survival prediction. MOGONET \citep{wang2021mogonet} introduces a GCN-based framework for integrating methylation, mRNA, and miRNA through a cross-omics discovery tensor and a view correlation discovery network. However, it identifies important features through computationally intensive ablation studies, and its graph construction relies on cosine similarity with a manually tuned edge-to-node ratio. 
MODILM \citep{zhong2023modilm} adopts a similar structure but replaces GCN with GAT, while CLCLSA \citep{Zhao2024CLCLSA:Data} addresses incomplete omics data using cross-omics autoencoders, contrastive learning, and attention mechanisms without explicitly modeling graph structures. SUPREME \citep{kesimoglu2023supreme} and MOGAT \citep{tanvir2024mogat} extend graph-based modeling to cancer subtype prediction and survival analysis by employing patient similarity networks derived from predefined metrics. 
DeepMoIC \citep{wu2024deepmoic} integrates three omics types using autoencoders, similarity network fusion, and a deep GCN with residual connections, while MoGCN \citep{li2022mogcn} focuses on breast cancer subtyping with graphs constructed via similarity network fusion. Although these models achieve promising results within their respective applications, they primarily rely on unsupervised or distance-based graphs and often lack built-in interpretability or robustness under class imbalance.

Despite recent advances, applying GNNs to multi-omics data remains challenging and relatively underexplored \citep{zhang2021graph, valous2024graph, paul2024systematic, ballard2024deep}. Existing methods exhibit three key limitations: 
(1) \textit{Inadequate modeling of high-dimensional, heterogeneous omics data}: To address the ``small $n$, large $p$'' issue, many methods apply simple filtering (e.g., removing low-variance or zero-valued features) or flatten multi-omics data through concatenation, which weakens modality-specific signals and ignores biological heterogeneity. Others rely on handcrafted similarity graphs (e.g., Pearson or cosine), which may miss nonlinear, task-specific relationships and are sensitive to arbitrary thresholding. 
(2) \textit{Vulnerability to class imbalance}: While class imbalance is common in biomedical datasets (e.g., rare diseases, early-stage cancer detections, or minority subtypes), it is rarely addressed during model development or evaluation. Many models emphasize overall accuracy, which can obscure poor minority-class performance and limit generalizability.  (3) \textit{Limited interpretability and biological insight}: Interpretability is essential in biomedical research for trust, transparency, and downstream discovery. However, most GNN-based models lack built-in interpretability and instead rely on computationally intensive post-hoc analyses or ablation studies that may be unstable. Furthermore, they typically treat all omics modalities equally, overlooking their relative contributions to prediction.

To address these limitations, we propose \textbf{MOTGNN}, a novel and interpretable graph-based framework for multi-omics data integration that combines supervised graph construction with hierarchical representation learning. Our main contributions are summarized as follows: (1) \textbf{Modality-specific modeling with supervised graph learning}: MOTGNN employs XGBoost to identify informative features for each omics type and uses the trained decision-tree structure to construct sparse, supervised, and modality-specific graphs. This approach preserves biologically meaningful interactions while reducing noise and redundancy. Omics-specific GNNs learn latent representations from each graph, which are then integrated through a deep feedforward network (DFN) to capture cross-omics interactions. (2) \textbf{Imbalance-robust architecture}: MOTGNN consistently outperforms existing models across all datasets, achieving 10-50\% improvement in F1-score on class-imbalanced real-world data. Its design mitigates overfitting to dominant classes and effectively detects signals from minority classes, addressing a critical limitation in biomedical classification tasks. (3) \textbf{Integrated interpretability at feature and omics levels}: MOTGNN provides end-to-end interpretability without requiring post-hoc analysis. It generates (i) feature-level importance scores to identify top-ranked biomarkers within each modality, and (ii) omics-level contribution scores to quantify the relative importance of each data type (e.g., methylation, mRNA, or miRNA), supporting both biological discovery and clinical insight.

\section{Methodology}
We summarize the key notations and their associated descriptions in Table~\ref{tab:notation}. A graph is denoted as $G(V, E)$, where $V$ represents the set of vertices (or nodes) and $E$ denotes the set of edges. The graph structure is encoded by its adjacency matrix $A$, where the $(i,j)$-th element is one if node $i$ is connected to node $j$, and zero otherwise. For undirected graphs, $A$ is symmetric. The adjacency matrix augmented with self-loops is denoted as $\tilde{A}$, where all diagonal elements are set to one to include self-connections.
\begin{table}[h]
\caption{Key notations and their descriptions used throughout the paper. }
\label{tab:notation}
\centering
\small
\begin{tabular}{@{}ll@{}}
\toprule
\textbf{Notation} & \textbf{Description} \\
\midrule
$G$ or $G(V, E)$ & Graph \\
$V$ & Set of vertices/nodes in $G$ \\
$E$ & Set of edges in $G$ \\
$A$ & Adjacency matrix of $G$ \\
$\tilde{A}$ & Adjacency matrix with self-loops \\
$X_i \in \mathbb{R}^{n \times p_i}$ & Omics data matrix for modality $i = 1, 2, 3$ \\
$n$ & Number of samples (rows in $X_i$) \\
$p_i$ &  Number of features in $X_i$  \\
$Y$ & Binary outcome variable \\
$X_i^* \in \mathbb{R}^{n \times p_i^*}$ & Dimension-reduced representation of $X_i$  \\
$p_i^*$  &  Reduced feature dimension for $X_i$ \\
$G_i$ or $G_i(V_i, E_i)$ &  Graph constructed for $X_i^*$ \\
$V_i$ & Set of vertices/nodes in $G_i$ \\
$E_i$ & Set of edges in $G_i$ \\
$Z_i$ & Embedding learned from $G_i$ via GNN \\
$Z$ & Concatenated representation of all $Z_i$ \\
$m_i$ & Edge-to-node ratio, i.e., $m_i = \frac{E_i}{p_i^*}$ \\
$RIG_i$ & Relative importance of graph $G_i$ \\
$\odot$ & Element-wise (Hadamard) product \\
$I(\cdot)$ & Indicator function\\
\bottomrule
\end{tabular}
\end{table}

\begin{figure*}[h]
  \centering
  \includegraphics[width=0.9\textwidth]{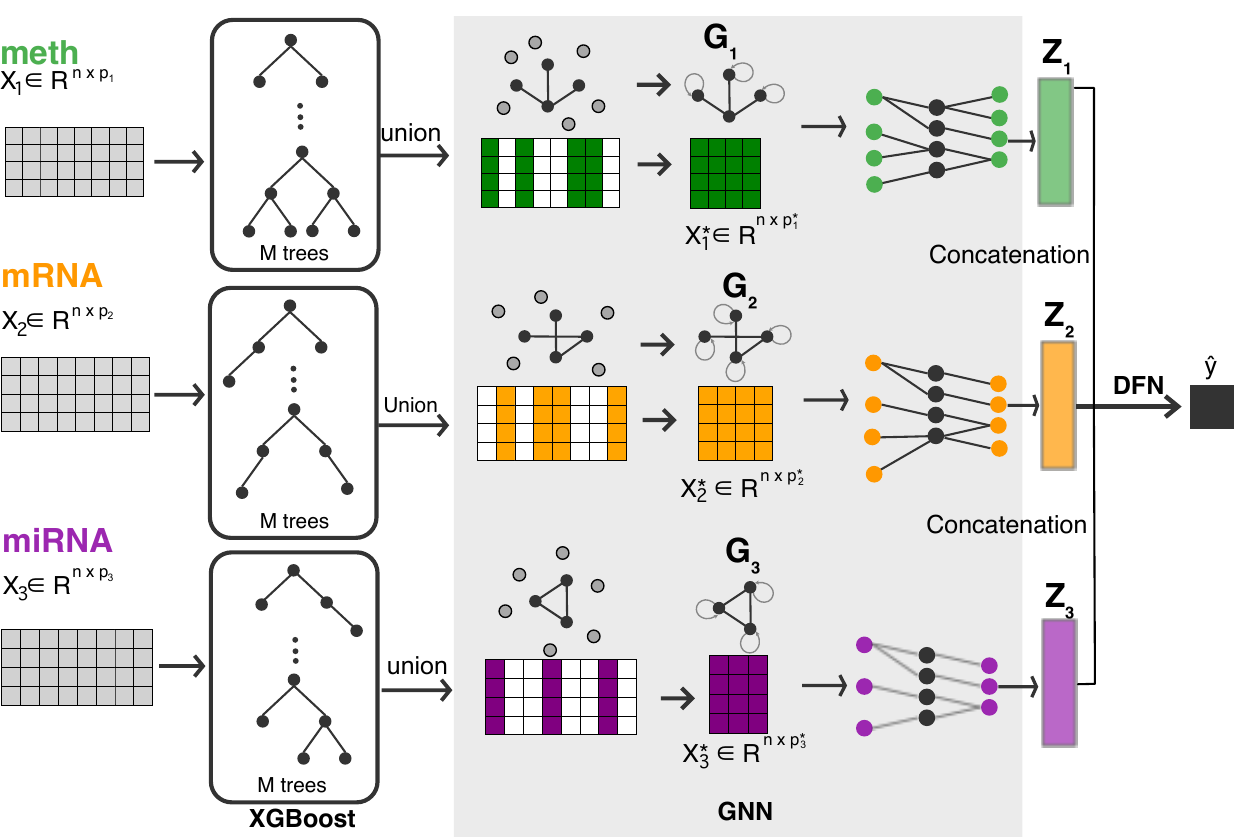}
  \caption{Overview of the proposed MOTGNN framework for multi-omics data integration and disease classification. The model comprises three key modules: (i) XGBoost for constructing omics-specific supervised feature graphs; (ii) graph neural network (GNN) for learning modality-specific embeddings by encoding each graph and its corresponding data matrix; and (iii) deep feedforward network (DFN) for integrating the learned embeddings and performing final classification.} \label{fig:MOTGNN}
\end{figure*}

\subsection{Proposed Model}
We propose MOTGNN, a \textbf{M}ulti-\textbf{O}mics integration framework with \textbf{T}ree-generated \textbf{G}raph \textbf{N}eural \textbf{N}etworks for binary disease classification. MOTGNN first leverages eXtreme Gradient Boosting (XGBoost) to select informative features in a supervised manner, as tree-based feature selection captures strong predictive signals. It then utilizes GNNs to learn omics-specific representations from graphs constructed using the structure of the trained XGBoost trees. This combination allows the model to extract both local and global patterns from each omics layer while maintaining biological interpretability. 

The overall framework of MOTGNN (Figure~\ref{fig:MOTGNN}) consists of three primary modules: (1) an \textbf{XGBoost module} for feature graph construction; (2) a \textbf{GNN module} to encode each omics-specific graph and its associated data matrix; and (3) a \textbf{DFN module} to integrate and classify the learned embeddings. This modular architecture allows MOTGNN to handle high-dimensional, heterogeneous multi-omics data efficiently, while providing interpretable insights at both the feature and omics levels. 
Algorithm~\ref{alg:MOTGNN} outlines the overall procedure of MOTGNN. 

\begin{algorithm}
\caption{MOTGNN} \label{alg:MOTGNN}
\begin{algorithmic}[1] 
\Require Omics datasets: $X_1 \in \mathbb{R}^{n \times p_1}$, $X_2 \in \mathbb{R}^{n \times p_2}$, $X_3 \in \mathbb{R}^{n \times p_3}$; Binary disease labels: $Y \in \mathbb{R}^{n \times 1}$
\Ensure Predicted labels $\hat{Y}$
\State Train three XGBoost models using $X_1$, $X_2$, and $X_3$ to predict $Y$
\State Construct omic-specific feature graphs: $G_1(V_1, E_1)$, $G_2(V_2, E_2)$, and $G_3(V_3, E_3)$ by taking the union of decision paths from all trees trained in Step 1, along with the corresponding dimension-reduced datasets: $X_1^* \in \mathbb{R}^{n \times p_1^*}$, $X_2^* \in \mathbb{R}^{n \times p_2^*}$, and $X_3^* \in \mathbb{R}^{n \times p_3^*}$
\For{each $(X_i^*, G_i)$, $i = 1, 2, 3$}
    \State Feed $X_i^*$ and $\tilde{A}_i$ (augmented adjacency matrix of $G_i$ with self-loops) into a GNN
    \State Obtain learned embedding $Z_i$ 
\EndFor  
\State Concatenate $Z_1$, $Z_2$, and $Z_3$ to form a unified representation $Z = [Z_1 | Z_2 | Z_3]$
\State Pass $Z$ into a DFN to perform final binary classification
\State \Return Predicted classification labels $\hat{Y}$
\end{algorithmic}
\end{algorithm}

\subsection{Graph Construction} \label{sec:dimensionReduction}
XGBoost \citep{chen2016xgboost} is a powerful ensemble method based on gradient-boosted decision trees. Compared with traditional gradient boosting machines, it introduces a regularized learning objective to reduce overfitting and improve generalization. In MOTGNN, we employ three XGBoost models, each trained independently to perform binary classification on the omics-specific datasets $X_{1}$, $X_{2}$, and $X_{3}$ together with the binary response $Y$. Unlike random forests, which generate independent trees, each XGBoost model generates $M$ sequential decision trees, where every tree improves upon the residuals of the previous ones. During training, each tree selects a subset of features for node splits. We collect the union of all features used across the $M$ trees and define these as the selected features. This yields a supervised, tree-guided feature selection process that reduces the original feature dimensions from $p_1$, $p_2$, and $p_3$ to $p_1^*$, $p_2^*$, and $p_3^*$, respectively, where $p_{i}^* < p_{i}$ and, in general, $p_1^*\neq p_2^*\neq p_3^*$. The reduced feature spaces preserve variables with higher predictive importance, facilitating downstream graph construction and neural modeling.

Inspired by forgeNet \citep{kong2020forgenet}, which constructs feature graphs based on the structure of ensemble trees, we adapt this idea to build omics-specific undirected graphs from XGBoost models. For each omics dataset, we treat each trained decision tree as a graph, where features involved in splits are represented as nodes and parent-child relationships form undirected edges. For example, consider $X_1 \in \mathbb{R}^{n \times p_1}$ and the binary outcome $Y$: if the XGBoost model generates $M$ sequential decision trees, each tree is viewed as an undirected graph $G^m = (V^m, E^m)$ for $m = 1, 2, \ldots, M$. We then take the union of all tree-level graphs to form an aggregated graph as $$G_{1}(V_1, E_1) = G_{1}(\bigcup_{m=1}^{M} V^{m}, \bigcup_{m=1}^{M} E^{m}),$$  
where the number of nodes is $|V_1| = p_1^*$, and $p_{1}^* < p_{1}$. Self-loops are added to the adjacency matrices to preserve self-information in GNN message passing. The similar procedure is used to generate $G_2$ and $G_3$ for the remaining two omics datasets.

\subsection{Graph Neural Network and Feature Fusion}
We use the graph-embedded deep feedforward network (GEDFN)  model \citep{kong2018graph} as the core GNN component used in MOTGNN. Unlike standard fully connected architectures, GEDFN directly incorporates the adjacency matrix into the weight connections between the input and the first hidden layer. This design allows the model to respect the feature-level relationships encoded in the graph while enforcing sparsity, which is a common assumption for omics-based feature graphs. The number of neurons in the first hidden layer is set equal to the input dimension. Given a reduced input data $X^* \in \mathbb{R}^{n \times p^*}$ and its augmented graph adjacency matrix $\tilde{A} \in \mathbb{R}^{p^* \times p^*}$, the first hidden layer $Z_1 \in \mathbb{R}^{n \times p^*}$ is computed as 
$$Z_1 = \sigma(X^*(W_{\text{in}}\odot \tilde{A}) + b_{\text{in}}),$$ 
where $W_{\text{in}} \in \mathbb{R}^{p^* \times p^*}$ and $b_{\text{in}} \in \mathbb{R}^{n \times p^*}$ are the weight and bias matrices, $\odot$ denotes element-wise (Hadamard) multiplication, and $\sigma(\cdot)$ is the activation function. This ensures that only graph-supported connections are active between the input features and the first hidden layer.
The design of GEDFN shares conceptual similarities with the graph attention network (GAT) formulation discussed in \citep{chen2023bridging}, where the GAT layer is represented as
$$Z_1 = \sigma((W_{\text{attention}}\odot A)X + b_{\text{in}}),$$
with $W_{\text{attention}} \in \mathbb{R}^{n \times n}$ and the nodes representing samples rather than features. In this formulation, GAT dynamically learns attention weights over neighbors. In GEDFN,  the graph structure constrains which connections are allowed, while the model still learns connection strengths. This yields a sparse and interpretable architecture tailored to omics data. 

After training three omics-specific GNNs, each on its own dimension-reduced input and constructed graph, we obtain three final hidden representations: $Z_1$, $Z_2$, and $Z_3$. These embeddings are concatenated to form a unified representation:
$$Z = [Z_1 | Z_2 | Z_3],$$ 
which is then passed to a DFN consisting of fully connected layers and a final softmax output layer to perform binary disease classification.

\begin{table*}[h]
\caption{Summary of three TCGA cancer datasets (COADREAD, LGG, and STAD), including sample sizes, class distributions, the number of original features, and the number of preprocessed features for each omics type: DNA methylation (meth), gene expression (mRNA), and microRNA expression (miRNA). The preprocessed data were obtained from the Broad GDAC Firehose.} 
\label{tab:datainfo}
\centering
\resizebox{\textwidth}{!} {%
\begin{tabular}{@{}llrrrr@{}}
\toprule
\textbf{Abbr.} & \textbf{Data Name} & \textbf{Samples} & \textbf{Class Balance} & \textbf{Original Features} & \textbf{Preprocessed Features} \\
& & & (0:1) & (meth:mRNA:miRNA)  & (meth:mRNA:miRNA) \\
\midrule
COADREAD & Colorectal and Rectal Adenocarcinoma & 332 & COAD:READ = 254:78 & 20,113:20,531:420 & 2,000:2,000:420\\
LGG & Low-Grade Glioma & 524 & Grade 2:Grade 3 = 255:269 & 20,531:20,114:548 & 2,000:2,000:548 \\ 
STAD & Stomach Adenocarcinoma & 371 & ADC:IAC = 205:166 & 20,101:20,531:507 & 2,000:2,000:507 \\ 
\bottomrule
\end{tabular}%
}
\end{table*}

\begin{figure*}[h]
  \centering
  \includegraphics[width=0.9\textwidth]{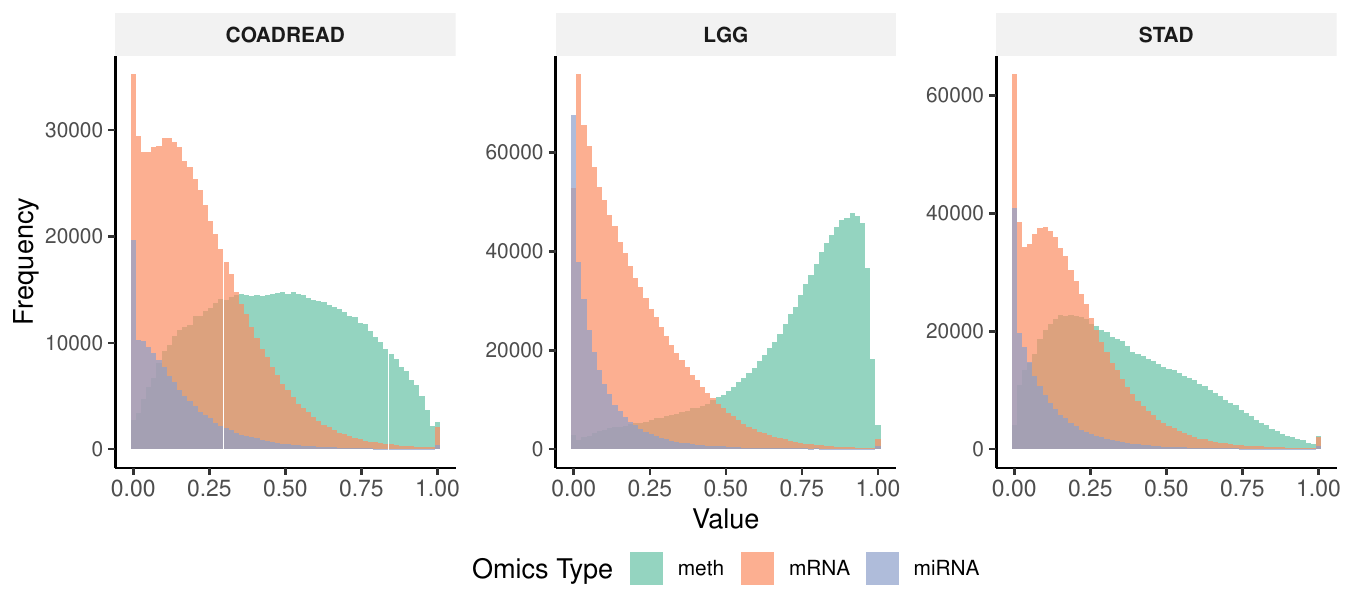}
  \caption{Distribution of DNA methylation, mRNA, and miRNA features across the preprocessed COADREAD, LGG, and STAD datasets. Each omics dataset was independently scaled to the range of [0, 1] using min-max normalization. The distinct distributional patterns highlight the heterogeneous characteristics of different omics types.} \label{fig:omicsHistogram}
\end{figure*}

\subsection{Interpretability on Features and Graphs}
To interpret model predictions and evaluate the contribution of both features and graphs, we adopt and extend the connection weights method \citep{olden2002illuminating}. Feature importance is quantified by summing the absolute values of weights connecting each input neuron to the first hidden layer, while accounting for adjacency-based sparsity in the feature graph. For feature $j$ in omics modality $X_i$, the importance score $IF_{j}^{i}$ is defined as:
$$IF_{j}^{i} = \sum_{u=1}^{p_{i}} \left|W_{ju}^{(\text{in})}I(\tilde{A}_{ju}^{i}=1)\right|, $$
where $i=1, 2, 3$, $W^{(\text{in})}$ is the input weight matrix and $I(\cdot)$ is the indicator function. The final importance score for feature $j$ across all omics modalities is obtained by summing its importance values: 
$$IF_{j} = IF_{j}^{1} + IF_{j}^{2} + IF_{j}^{3}.$$
After training the model, each input feature is assigned an importance score that can be used to rank and identify top biomarkers for downstream biological investigation.
To assess the relative importance of each omics-specific graph, we compute the $L_1$ norm of the weight matrix connecting the GNN’s last hidden representations ($Z_i$) to the DFN layer. The relative graph importance ($RIG_i$) is defined as:
\begin{eqnarray*}
RIG_{i} = \frac{\|W_{Z_i\leftrightarrow f}\|_{1}}{\sum_{i=1}^3\|W_{Z_i\leftrightarrow f}\|_{1}},
 \end{eqnarray*}
where $W_{Z_i\leftrightarrow f}$ denotes the weight matrix connecting $Z_i$ to the DFN, and $\|\cdot\|_{1}$ is the $L_1$ norm (the sum of absolute values). The total relative importance across all three graphs sums to one.

\subsection{Technical Details}
We implement the XGBoost module in MOTGNN using the \textsc{xgboost} Python library and tune the number of estimators (i.e., the number of trees). The GNN and DFN modules are implemented in the \textsc{Tensorflow} library \citep{abadi2016tensorflow}. Model optimization uses the Adam optimizer \citep{kingma2014adam} with mini-batch training. The nonlinear activation function uses the Rectified Linear Unit (ReLU) function: $\sigma(x) = max(0, x)$ \citep{nair2010rectified}. For classification, the final output layer uses a softmax function, and the loss is computed using binary cross-entropy: $$L(Y, \hat{Y}) = -\frac{1}{n} \sum_{i=1}^n \left[ y_i \ln(\hat{y}_i) + (1 - y_i) \ln(1 - \hat{y}_i) \right],$$ 
where $y_i \in \{0, 1\}$ is the ground truth label, and $\hat{y}_i$ is the predicted probability of the positive class. 

We tune hyperparameters across three main categories: (1) architectural (number of layers and hidden neurons), (2) training-related (learning rate, batch size, and epochs), and (3) regularization-related (dropout and $L_2$ penalty). 
The $L_2$ regularization term is expressed as
$L_{\text{reg}} = \lambda \sum_{j} w_j^2$,
where $\lambda$ is the regularization coefficient and $w_j$ denotes model weights. 
To avoid overfitting, we apply various strategies including dropout, batch normalization, $L_2$ regularization, and early stopping. Data are split into training, validation, and test sets in a 60\%:20\%:20\% ratio, with stratified sampling used to address class imbalance. We perform 20 independent and reproducible train-validation-test splits. Hyperparameters are tuned via a grid search within a reasonable space and informed by prior experience. 

\section{Real Data Applications}

\subsection{Data Description}
We evaluate our method on three real-world cancer datasets from The Cancer Genome Atlas (TCGA) \citep{weinstein2013cancer}: COADREAD, STAD, and LGG. The processed multi-omics data and label annotations were obtained from the Broad GDAC Firehose (\url{https://gdac.broadinstitute.org/}). Each dataset includes matched DNA methylation ($X_{1}$), mRNA expression ($X_{2}$), and miRNA expression ($X_{3}$), along with class labels ($Y$). Only samples with complete measurements across all three omics types were included in our analysis. The COADREAD dataset combines \textbf{CO}lorectal \textbf{AD}enocarcinoma (COAD) and \textbf{RE}ctal \textbf{AD}enocarcinoma (READ) into a binary classification task, with 254 COAD and 78 READ samples.
The LGG dataset involves grade classification for low-grade gliomas, with 255 Grade 2 and 269 Grade 3 samples. 
The STAD dataset targets the classification of \textbf{ST}omach \textbf{AD}enocarcinoma subtypes, with 205 adenocarcinoma (ADC) and 166 intestinal adenocarcinoma (IAC) samples. Among the three datasets, COADREAD exhibits the most imbalanced class distribution. 

Table~\ref{tab:datainfo} summarizes the datasets in terms of sample sizes, class distributions, and the number of original and preprocessed features for each omics type. Data preprocessing included standard quality control steps to remove noisy and redundant features, such as filtering out zero-expression signals and low-variance features, as well as adjusting for multiple comparisons \citep{Lu2023MultiomicsSubtypes}. Each omics dataset was independently normalized to the range [0, 1] using min-max scaling. 
The preprocessed data shared in \citep{Lu2023MultiomicsSubtypes} contained minor inconsistencies in sample sizes across omics types. For COADREAD, methylation and mRNA data each included 332 samples, while miRNA had 337. We removed the five extra miRNA samples to ensure alignment. Similarly, for STAD, methylation and mRNA data each had 371 samples, whereas miRNA had 372; we excluded one redundant miRNA sample for consistency. 
The histograms in Figure~\ref{fig:omicsHistogram} show the feature distributions across three omics types (methylation, mRNA, and miRNA) for COADREAD, LGG, and STAD. The distinct patterns among omics types highlight the inherent heterogeneity and complementary nature of each modality. These differences across datasets help evaluate the generalizability of MOTGNN. 

\subsection{Comparison and Evaluation Metrics}
We evaluate the performance of our proposed model, MOTGNN, which by default employs XGBoost to generate three graphs (one per omics type) and uses GEDFN as the main graph neural network (GNN) component. The model is compared with \textit{four baseline machine learning and deep learning models}: XGBoost, random forest (RF), deep feedforward network (DFN), and graph convolutional network (GCN).  
To assess the contribution of individual architectural components, we also design \textit{two ablation variants} of MOTGNN: (1) $\mathrm{MOTGNN}_{\text{gcn}}$ replaces the GEDFN module with a standard GCN, and (2) $\mathrm{MOTGNN}_{\text{rf}}$ uses RF instead of XGBoost for graph generation. These ablation models share the same overall architecture as MOTGNN but differ only in the modified component. 

The number of trees for both XGBoost and RF is set to 100. The DFN and GCN models each have two hidden layers with 64 neurons per layer. 
For XGBoost, RF, DFN, and GCN, the input consists of the column-wise concatenation of the three preprocessed omics data types (methylation, mRNA, and miRNA). In contrast, $\mathrm{MOTGNN}_{\text{gcn}}$ and $\mathrm{MOTGNN}_{\text{rf}}$ process each omics modality separately, using three parallel GNNs, one for each omics type. The graphs used in GCN and $\mathrm{MOTGNN}_{\text{gcn}}$ are generated by XGBoost, while $\mathrm{MOTGNN}_{\text{rf}}$ are generated by RF. 
Table~\ref{tab:hyperparams} summarizes the hyperparameter tuning ranges and selected values (optimal choices in bold) of MOTGNN. 

We report three standard evaluation metrics: accuracy, ROC-AUC, and F1-score. Accuracy measures the proportion of correctly classified samples. ROC-AUC (area under the receiver operating characteristic curve) quantifies the model’s ability to distinguish between classes. F1-score is defined as 
$$F1 = \frac{2\times \text{precision} \times \text{recall}}{\text{precision} + \text{recall}},$$
where precision refers to the positive predictive value and recall corresponds to sensitivity (true positive rate). As the harmonic mean of precision and recall, the F1-score provides a balanced assessment under imbalanced class distributions, mitigating bias toward the majority class. All computations were conducted on the Falcon supercomputer \citep{falcon2022}, using four CPUs and 20 GB of memory per job.
\begin{table}[h]
\caption{Hyperparameter tuning ranges for the XGBoost and neural network components in MOTGNN across the COADREAD, LGG, and STAD datasets. Optimal values are highlighted in bold. }
\label{tab:hyperparams}
\centering
\begin{tabular}{lr}
\toprule
\textbf{Hyperparameter} & \textbf{Tuning Range} \\
\midrule
\multicolumn{2}{l}{\textit{XGBoost Parameters}} \\
Number of Trees & \textbf{100}, 1000 \\
\midrule
\multicolumn{2}{l}{\textit{Neural Network Architecture}} \\
Depth (hidden layers) & 1, \textbf{2}, 3 \\
Width (neurons per layer) & 32, \textbf{64}, 128 \\
Activation Function & \textbf{ReLU}, leakyReLU \\
\midrule
\multicolumn{2}{l}{\textit{Training Parameters}} \\
Learning Rate & 0.001, 0.0005, \textbf{0.0001} \\
Batch Size & 8, \textbf{16}, 32 \\
Training Epochs & 200, 300, \textbf{500} \\
\midrule
\multicolumn{2}{l}{\textit{Regularization and Early Stopping}} \\
Dropout Rate & 0.2, 0.3, \textbf{0.5} \\
$L_2$ Regularization Coefficient ($\lambda$) & \textbf{0.01} \\
Early Stopping Patience & 5, \textbf{10} \\
Early Stopping Min Delta & \textbf{0.001}, 0.005 \\
\bottomrule
\end{tabular}
\end{table}

\begin{figure}[h]
  \centering
  \includegraphics[width=\columnwidth]{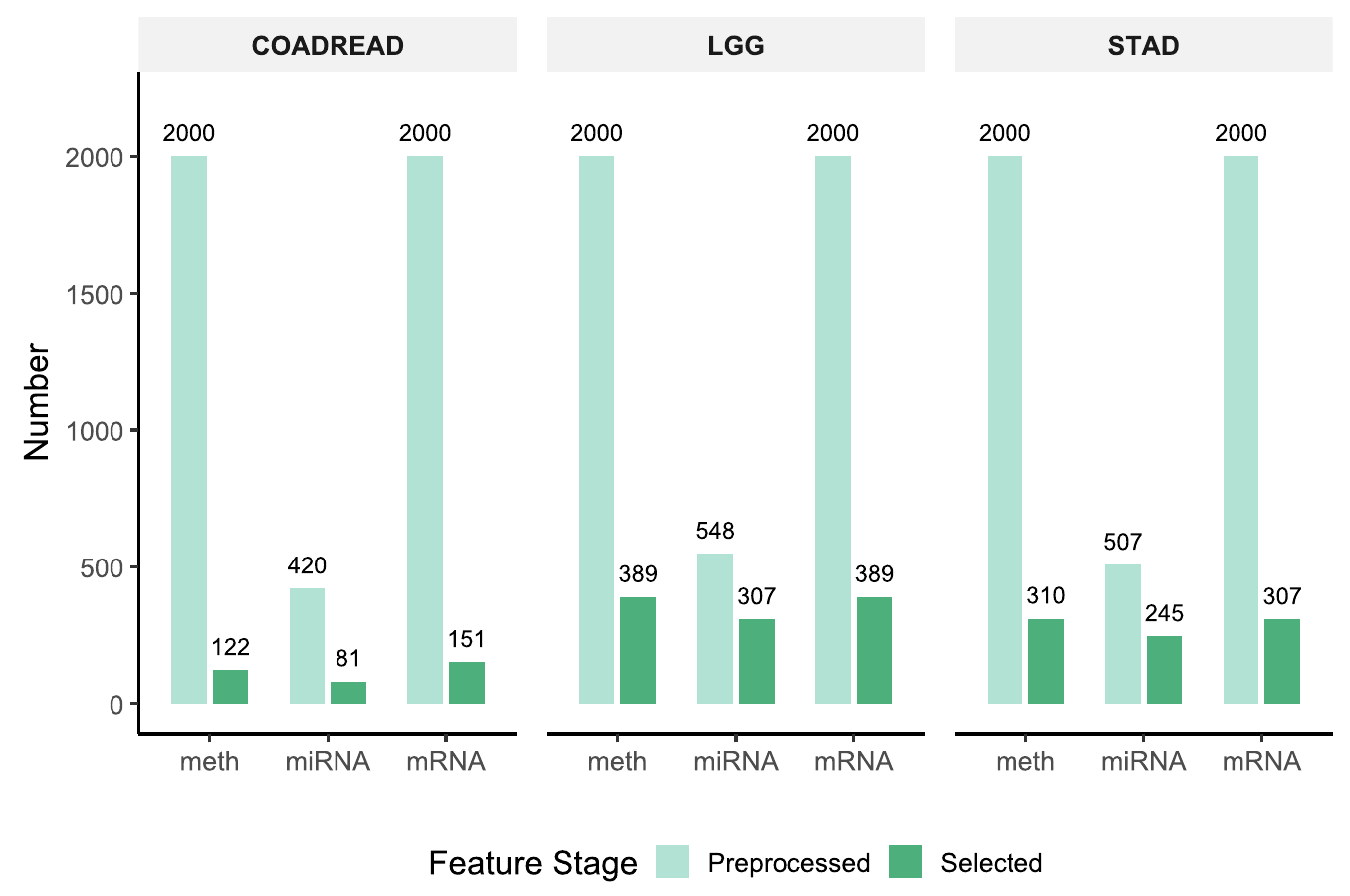}
  \caption{Feature dimensions before and after XGBoost-based selection on COADREAD, LGG, and STAD datasets. The bar plots compare preprocessed feature dimensions (pre-selection) with reduced dimensions (post-selection), showing substantial dimensionality reduction across all omics types.
  } \label{fig:featureselection}
\end{figure}

\begin{table}[h]
\caption{Structural properties of sparse graphs constructed by MOTGNN for COADREAD, LGG, and STAD datasets. $E_i$ denotes the number of edges (including self-loops) in graph $G_i$, and $m_i$ represents the edge-to-node ratio, $m_i = E_i/p_i^*$.}
\label{tab:graphstats}
\centering
\small
\begin{tabular}{@{}lrr@{}}
\toprule
\textbf{Dataset} & \multicolumn{1}{c}{\textbf{Edges}} & \multicolumn{1}{c}{\textbf{Edge/Node}} \\
& ($E_1$:$E_2$:$E_3$) & ($m_1$:$m_2$:$m_3$) \\
\midrule
COADREAD & 260:318:190 & 2.13:2.11:2.35 \\
LGG & 880:849:856 & 2.26:2.18:2.79 \\
STAD & 674:646:639 & 2.17:2.10:2.61 \\
\bottomrule
\end{tabular}
\end{table}

\begin{table*}[h]
\caption{Classification performance comparison of MOTGNN and other models on COADREAD, LGG, and STAD datasets. Results are reported as mean ± standard deviation [95\% confidence interval] over 20 independent runs. Baseline models include XGBoost, RF (random forest), DFN (deep feedforward network), and GCN (graph convolutional network). 
$\mathrm{MOTGNN}_{\text{gcn}}$ and $\mathrm{MOTGNN}_{\text{rf}}$ represent MOTGNN variants that respectively replace GEDFN with GCN or use RF instead of XGBoost for graph generation. Bold values indicate the best performance.
}
\label{tab:classification}
\centering
\begin{tabular}{@{}llrrr@{}}
\toprule
\textbf{Dataset} & \textbf{Model} & \textbf{Accuracy} & \textbf{ROC-AUC} & \textbf{F1-Score} \\
\midrule
\multirow{5}{*}{COADREAD} 
& XGBoost & 0.910 ± 0.034 [0.894, 0.926] & 0.964 ± 0.022 [0.953, 0.975] & 0.779 ± 0.094 [0.735, 0.823] \\
& RF & 0.801 ± 0.025 [0.789, 0.813] & 0.891 ± 0.038 [0.873, 0.909] & 0.334 ± 0.119 [0.276, 0.392] \\
& DFN & 0.873 ± 0.043 [0.853, 0.893] & 0.926 ± 0.047 [0.904, 0.948] & 0.741 ± 0.092 [0.696, 0.786] \\
& GCN & 0.871 ± 0.051 [0.847, 0.895] & 0.920 ± 0.049 [0.897, 0.942] & 0.735 ±  0.107 [0.685, 0.785]\\
& $\mathrm{MOTGNN}_{\text{gcn}}$ & 0.895 ± 0.039 [0.877, 0.913] & 0.928 ± 0.034 [0.912, 0.944] & 0.783 ± 0.078 [0.746, 0.820] \\
& $\mathrm{MOTGNN}_{\text{rf}}$  & 0.933 ± 0.028 [0.920, 0.946] & 0.967 ± 0.026 [0.955, 0.979] & 0.859 ± 0.061 [0.830, 0.888] \\
& MOTGNN & \textbf{0.939 ± 0.031 [0.925, 0.953]} & \textbf{0.969 ± 0.023 [0.958, 0.980]} & \textbf{0.872 ± 0.064 [0.842, 0.902]} \\
\midrule
\multirow{5}{*}{LGG}
& XGBoost & 0.663 ± 0.041 [0.644, 0.682] & 0.718 ± 0.032 [0.703, 0.733] & 0.661 ± 0.041 [0.642, 0.680] \\
& RF & 0.691 ± 0.051 [0.667, 0.715] & 0.764 ± 0.044 [0.743, 0.785] & 0.668 ± 0.063 [0.638, 0.698] \\
& DFN & 0.691 ± 0.053 [0.666, 0.716] & 0.759 ± 0.049 [0.736, 0.782] & 0.694 ± 0.052 [0.670, 0.718] \\
& GCN &  0.692 ± 0.051 [0.668, 0.716] &  0.763 ± 0.049 [0.740, 0.785] & 0.693 ± 0.057 [0.666, 0.719] \\
& $\mathrm{MOTGNN}_{\text{gcn}}$ & 0.696 ± 0.050 [0.673, 0.719] & 0.764 ± 0.047 [0.742, 0.786] & 0.692 ± 0.052 [0.668, 0.716] \\
& $\mathrm{MOTGNN}_{\text{rf}}$  & 0.706 ± 0.030 [0.692, 0.720] & 0.777 ± 0.038 [0.759, 0.795] & 0.703 ± 0.036 [0.686, 0.720] \\
& MOTGNN & \textbf{0.711 ± 0.038 [0.693, 0.729]} & \textbf{0.777 ± 0.046 [0.756, 0.798]} & \textbf{0.710 ± 0.039 [0.692, 0.728]} \\
\midrule
\multirow{5}{*}{STAD}
& XGBoost & 0.597 ± 0.061 [0.569, 0.625] & 0.635 ± 0.070 [0.603, 0.667] & 0.542 ± 0.067 [0.511, 0.573] \\
& RF & 0.628 ± 0.040 [0.609, 0.647] & 0.681 ± 0.052 [0.657, 0.705] & 0.559 ± 0.053 [0.534, 0.584] \\
& DFN & 0.635 ± 0.035 [0.619, 0.651] & 0.691 ± 0.049 [0.668, 0.714] & 0.595 ± 0.046 [0.573, 0.617] \\
& GCN &  0.640 ± 0.047 [0.618, 0.662] &  0.694 ± 0.045 [0.673, 0.715] & 0.603 ± 0.053 [0.578, 0.628] \\
& $\mathrm{MOTGNN}_{\text{gcn}}$ & 0.653 ± 0.062 [0.624, 0.682] & 0.686 ± 0.057 [0.659, 0.713] & 0.613 ± 0.076 [0.577, 0.649] \\
& $\mathrm{MOTGNN}_{\text{rf}}$  & 0.657 ± 0.048 [0.635, 0.679] & 0.710 ± 0.049 [0.687, 0.733] & 0.616 ± 0.058 [0.589, 0.643] \\
& MOTGNN & \textbf{0.664 ± 0.060 [0.636, 0.692]} & \textbf{0.712 ± 0.069 [0.680, 0.744]} & \textbf{0.627 ± 0.063 [0.597, 0.657]} \\
\bottomrule
\end{tabular}
\end{table*}

\begin{figure*}[h]
  \centering
  \includegraphics[width=\textwidth]{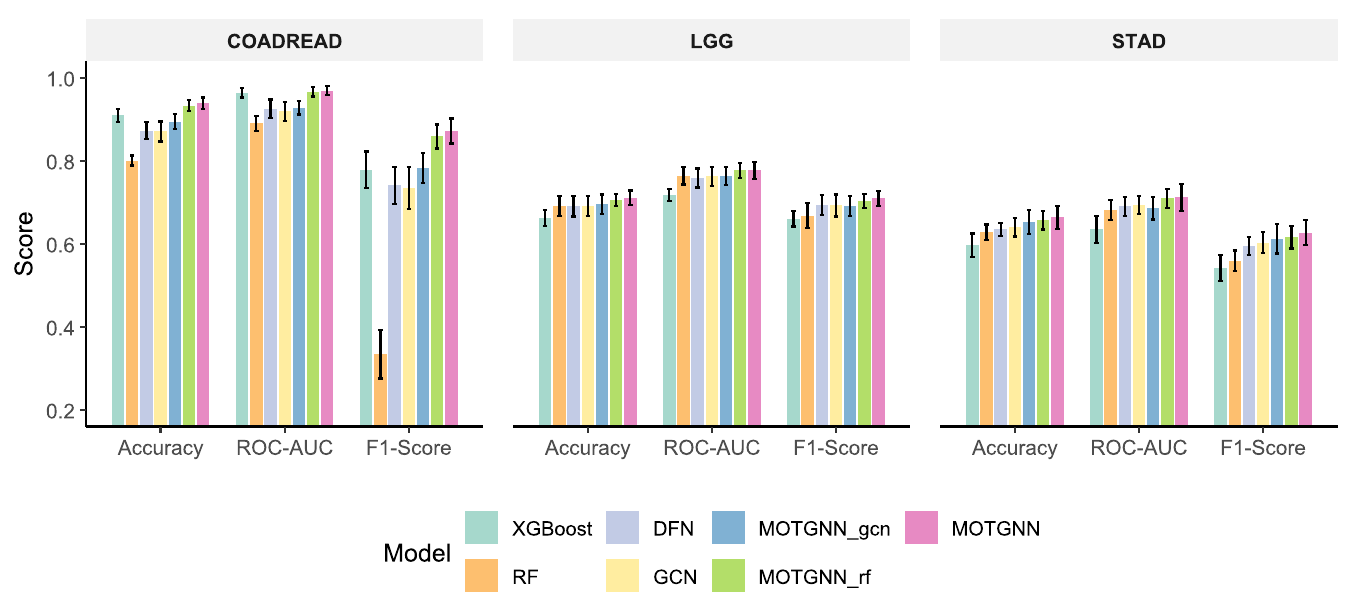}
  \caption{Comparison of classification performance across COADREAD, LGG, and STAD datasets. Bars show mean scores, and error bars represent 95\% confidence intervals over 20 independent runs. MOTGNN consistently achieves the highest scores across accuracy, ROC-AUC, and F1 metrics.} \label{fig:classification}
\end{figure*}

\begin{figure*}[h]
  \centering
  \includegraphics[width=0.85\textwidth]{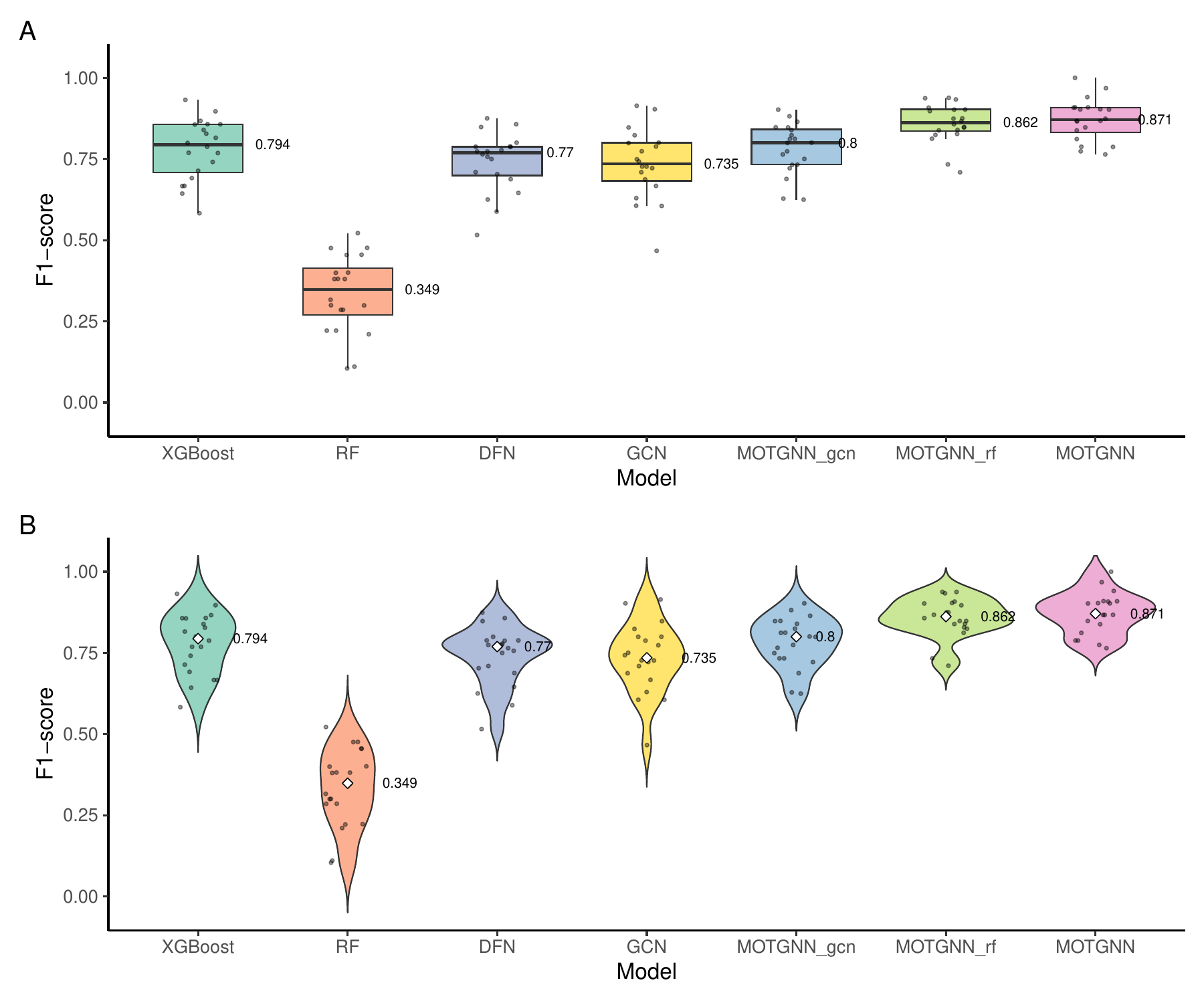}
  \caption{F1-score comparison across 20 independent runs on the imbalanced COADREAD dataset (class ratio 254:78). \textbf{(A)} Box plots display the median (labeled), interquartile range, and outliers. \textbf{(B)} Violin plots illustrate the distribution of F1-scores with labeled medians. MOTGNN achieves the highest median score with the lowest variability, demonstrating stable and robust performance under class imbalance.} \label{fig:BoxandViolinPlot}
\end{figure*}

\begin{figure}[h]
  \centering
\includegraphics[width=\columnwidth]{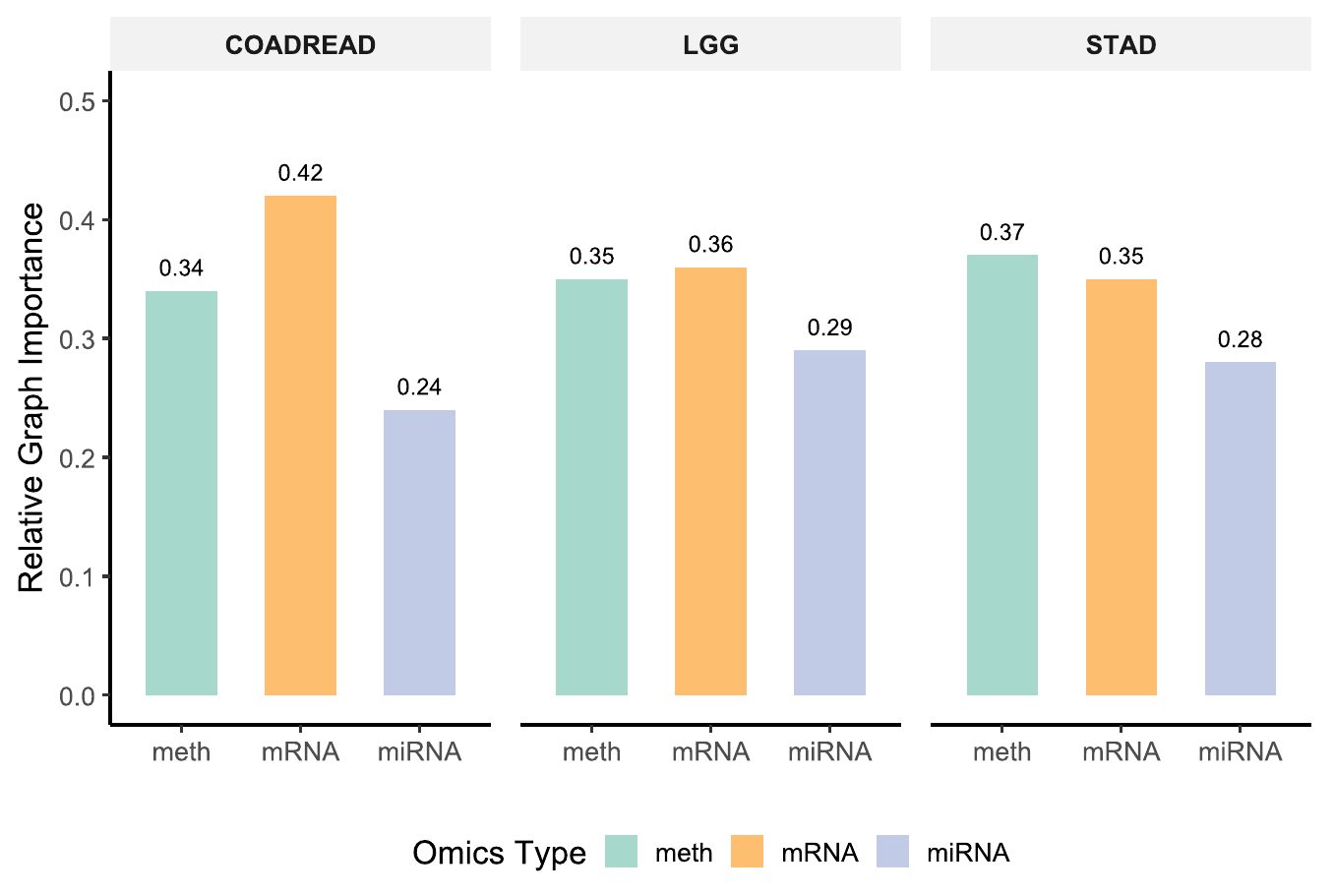}
  \caption{Relative importance of feature graphs from different omics types (methylation, mRNA, and miRNA) in MOTGNN across COADREAD, LGG, and STAD datasets. Bar heights indicate the contribution of each omics layer to the overall classification performance, revealing dataset-specific biological dominance patterns. } \label{fig:graphImportance}
\end{figure}

\section{Results and Discussion}

\noindent \textbf{Feature Selection and Graph Properties}. Figure~\ref{fig:featureselection} presents the feature dimensions before and after applying XGBoost within the MOTGNN framework. The reduced omics matrices $X_{1}^*$, $X_{2}^*$, and $X_{3}^*$ represent the selected feature subsets for DNA methylation, mRNA, and miRNA, respectively.  
The reduced feature sizes ($p_{i}^*$) vary across omics types due to the nature of XGBoost’s supervised feature selection: the final selected set is the union of features used across all $M$ decision trees, as described in Section~\ref{sec:dimensionReduction}. Table~\ref{tab:graphstats} summarizes the structural properties of the three constructed feature graphs ($G_1$, $G_2$, $G_3$), where $E_i$ denotes the set of edges for Graph $G_i$ and the edge-to-node ratio is defined as $m_i = \frac{E_i}{p_i^*}$ for $i = 1, 2, 3$. 

\noindent \textbf{Classification Performance}. Table~\ref{tab:classification} presents the classification results averaged over 20 independent train-validation-test splits. Overall, MOTGNN consistently outperforms all baseline models (XGBoost, RF, DFN, and GCN) across all evaluation metrics (accuracy, ROC-AUC, and F1-score) on all datasets (COADREAD, LGG, and STAD). In COADREAD, MOTGNN achieves an accuracy of 93.9\%, surpassing XGBoost (91.0\%), RF (80.1\%), DFN (87.3\%), and GCN (87.1\%). The improvement is even more pronounced in F1-score, where MOTGNN attains 87.2\%, marking a 9.3\% absolute improvement over XGBoost (77.9\%) and a substantial 53.8\% advantage over RF (33.4\%). Given COADREAD’s imbalanced class distribution (254:78), these gains highlight MOTGNN’s robustness under skewed data conditions, where traditional ensemble models degrade substantially.
The two ablation variants, $\mathrm{MOTGNN}_{\text{gcn}}$ and $\mathrm{MOTGNN}_{\text{rf}}$, both outperform their respective base counterparts (GCN and RF), confirming the benefits of the modality-specific multi-graph design. However, the full MOTGNN remains superior, demonstrating that combining XGBoost-based graph construction with the GEDFN module provides complementary advantages for capturing nonlinear and omics-specific dependencies. 

Across all datasets, MOTGNN maintains the highest ROC-AUC values ($\geq$ 96.9\% for COADREAD and  $\geq$ 3\% higher than the best baseline in LGG and STAD). While XGBoost performs well in COADREAD (96.4\%), its performance declines on LGG (71.8\%) and STAD (63.5\%), indicating limited generalizability. DFN exhibits competitive baseline performance, particularly in STAD (69.1\% ROC-AUC), but lacks the ability to capture inter-feature relationships effectively. GCN achieves moderate gains by leveraging graph structure, but it constructs a single concatenated graph across omics, which may weaken omics-specific signals and introduce noise from irrelevant cross-omics interactions.  By contrast, MOTGNN learns individual omics-specific graphs and integrates them through interpretable GNN modules, achieving consistent and robust performance across all datasets.  These results confirm its effectiveness for complex multi-omics classification tasks. 

Figure~\ref{fig:classification} compares model performance across all datasets and evaluation metrics, highlighting MOTGNN’s consistent advantage over both baseline and ablation models. Although some confidence intervals slightly overlap, MOTGNN consistently attains the highest mean scores across accuracy, ROC-AUC, and F1-score. The largest performance gaps appear in F1-scores for COADREAD, where MOTGNN (87.2\%) far exceeds RF (33.4\%) and outperforms all other methods, including $\mathrm{MOTGNN}{\text{gcn}}$ and $\mathrm{MOTGNN}{\text{rf}}$. The figure also shows that, despite XGBoost’s strong performance in COADREAD (91\% accuracy), its accuracy drops markedly in LGG and STAD, underscoring limited robustness across datasets.
Figure~\ref{fig:BoxandViolinPlot} further demonstrates model robustness on the imbalanced COADREAD dataset (class ratio 254:78) by comparing F1-scores over 20 independent runs. MOTGNN achieves the highest median F1-score (0.871), outperforming XGBoost (0.794), DFN (0.771), GCN (0.735), RF (0.349), and both ablation variants $\mathrm{MOTGNN}{\text{gcn}}$ (0.862) and $\mathrm{MOTGNN}{\text{rf}}$ (0.800). The violin plots reveal a compact distribution around MOTGNN’s median, indicating low variability and stable performance. In contrast, XGBoost, DFN, and GCN display wider interquartile ranges, while RF shows low performance and high variance. The ablation variants outperform their respective baselines but remain below the full MOTGNN, reinforcing that both the GEDFN backbone and XGBoost-based graph generation contribute to the model’s overall strength. These results underscore the robustness of MOTGNN under class imbalance, a key challenge in biomedical data analysis.

\noindent \textbf{Graph Importance and Omics-Level Interpretability}. Figure~\ref{fig:graphImportance} displays the relative graph importance for the three constructed graphs. The observed edge-to-node ratios (ranging from approximately 2.1 to 2.8, as shown in Table~\ref{tab:graphstats}) indicate that all graphs are sparse. Among them, the miRNA graph ($G_3$) shows the highest edge density (2.35-2.79) but the lowest relative importance (24-29\%), suggesting that it contributes less to classification performance despite being more connected. 
In COADREAD, the mRNA graph ($G_2$) contributes most strongly, with an importance of roughly 42\%. In LGG and STAD, methylation ($G_1$) and mRNA ($G_2$) contribute comparably, ranging from 35\% to 37\%. The consistently high methylation importance across datasets (approximately 34-37\%) underscores its robustness as a discriminative biomarker, whereas the variability of mRNA importance reflects tissue-specific gene expression dynamics. These results demonstrate that MOTGNN effectively identifies the most informative omics modalities while maintaining computational efficiency through sparse graph construction. The resulting edge counts remain tractable, supporting scalability to larger or clinically oriented datasets.

\noindent \textbf{Biomarker Selection and Feature-Level Interpretability}. Table~\ref{tab:topBiomarkers} lists the top 30 most important biomarkers identified by MOTGNN for the COADREAD dataset, illustrating the feature-level interpretability of our framework. COADREAD encompasses both colorectal and rectal adenocarcinomas, which are often analyzed jointly because of their overlapping molecular signatures and clinical profiles. This dataset was selected for interpretability analysis due to its well-characterized molecular landscape and the strong classification performance achieved by MOTGNN.  The same feature-ranking procedure can also be applied to the other two cancer types: STAD (Stomach Adenocarcinoma) and LGG (Lower Grade Glioma). Among the top-ranked features identified for COADREAD, Secreted Frizzled-Related Protein 4 (SFRP4) stands out as a well-documented biomarker associated with colorectal cancer \citep{huang2010sfrp4, frank2010single, liu2020ezh2, yu2022sfrp4}. Multiple studies have reported that SFRP4 is overexpressed in tumor tissues, and its promoter hypermethylation occurs as an early event in cancer progression. Functionally, SFRP4 acts as a modulator of Wnt signaling, thereby influencing cell proliferation and tumor growth.
While the present work focuses on computational discovery through feature-importance analysis, these findings highlight MOTGNN’s potential for guiding downstream biological validation. The selected biomarkers serve as promising candidates for further investigation and may ultimately contribute to discovering novel diagnostic or therapeutic strategies for complex diseases.

\begin{table}[h]
\caption{Top 30 biomarkers identified by MOTGNN for the COADREAD dataset (Colorectal and Rectal Adenocarcinoma), grouped by omics type: DNA methylation (meth), gene expression (mRNA), and microRNA expression (miRNA). The listed biomarkers represent the biomarkers with the highest feature-importance scores derived from MOTGNN.}
\label{tab:topBiomarkers}
\centering
\small
\begin{tabular}{@{}ll@{}}
\toprule
\textbf{Omics Type} & \textbf{Biomarkers} \\
\midrule
\multirow{2}{*}{meth} 
&  LRRC37A2, C9orf70, SNORD98, SFRP4, ELAVL2\\
&  HOXD9, HOXC6, DENND2C, MIR519B, DDIT4L\\
\midrule
\multirow{4}{*}{mRNA} 
&  MUC12|10071, KDELR2|11014, H3F3A|3020\\
&  RPL21|6144, AKT1|207, HOXB13|10481 \\
&  ILVBL|10994, MAEA|10296, PRAC|84366\\
& H3F3C|440093 \\
\midrule
\multirow{3}{*}{miRNA} 
& hsa-let-7f-2, hsa-let-7g, hsa-mir-10b, hsa-mir-1201\\
& hsa-mir-1270, hsa-mir-1274b, hsa-mir-191\\
& hsa-mir-206, hsa-mir-30c-2, hsa-mir-425\\
\bottomrule
\end{tabular}%
\end{table}

Table~\ref{tab:computationTime} summarizes the computational trade-offs associated with these performance improvements. Among all models, MOTGNN requires the longest training time, averaging 1.75 minutes for COADREAD and up to 2.39 minutes for LGG, compared to GCN's 0.59-0.94 minutes, DFN's 0.3-0.39 minutes, XGBoost's 0.16-0.25 minutes, and RF's remarkably fast 0.05 minutes per run. Despite this additional cost, the runtime remains practical for real-world applications, especially considering the analysis was conducted using only four CPUs per job. The longer runtime primarily reflects the additional computational steps required for supervised graph construction and hierarchical representation learning. These added computations are justified by MOTGNN's superior and stable performance across diverse datasets and evaluation metrics, demonstrating a favorable balance between efficiency and predictive accuracy.
\begin{table}[h]
\caption{Computation time comparison of MOTGNN and baseline models (XGBoost, RF, DFN, and GCN) across COADREAD, LGG, and STAD datasets. Each experiment was conducted using only four CPUs without GPU acceleration, and the reported times are averaged over 20 independent runs.}
\label{tab:computationTime}
\centering
\small
\begin{tabular}{@{}llr@{}}
\toprule
\textbf{Data} & \textbf{Model} & \textbf{Time (min)} \\
\midrule
\multirow{5}{*}{COADREAD} 
& XGBoost & 0.25 \\
& RF & 0.045 \\
& DFN & 0.385 \\
& GCN & 0.585 \\
& MOTGNN & 1.75 \\
\midrule
\multirow{5}{*}{LGG} 
& XGBoost & 0.195\\
& RF &  0.05 \\
& DFN & 0.36 \\
& GCN &  0.94 \\
& MOTGNN & 2.385\\
\midrule
\multirow{5}{*}{STAD} 
& XGBoost & 0.165\\
& RF & 0.05 \\
& DFN & 0.3 \\
& GCN &  0.805 \\
& MOTGNN & 1.45\\
\bottomrule
\end{tabular}%
\end{table}

\section{Conclusion}
We present MOTGNN, a novel graph neural network framework for multi-omics data integration and disease classification. By leveraging XGBoost for omics-specific supervised graph construction and modality-specific graph neural networks for latent representation learning, MOTGNN effectively captures complex relationships among features that are often overlooked by traditional methods. Applied to real-world cancer datasets, MOTGNN consistently outperforms baseline models across evaluation metrics, demonstrating its strength in integrating heterogeneous omics data and improving predictive performance. In particular, MOTGNN maintains strong robustness under class imbalance, a common challenge in biomedical data analysis. In addition to its predictive power, MOTGNN provides interpretability at both the feature and omics levels. Feature-level importance scores enable the identification of top-ranked biomarkers, which may serve as promising candidates for further biological investigation. Meanwhile, omics-level contributions reveal the relative informativeness of each data modality (e.g., methylation, mRNA, or miRNA), offering valuable insight into their roles across different disease types. Together, these interpretability results show that MOTGNN not only enhances prediction performance but also yields biologically meaningful insights that can guide downstream validation and hypothesis generation. 

Future extensions of the MOTGNN framework may include incorporating additional omics modalities, expanding to other cancer types and non-cancer diseases, or adapting the architecture for more complex prediction tasks, such as multi-class classification or survival analysis. Further enhancing interpretability and computational scalability will also be important directions, particularly to better support clinical and biological decision-making. In the long term, MOTGNN’s modular design and interpretable learning process provide a flexible foundation for advancing integrative multi-omics research and enabling precision medicine applications.

\section*{Data and Code Availability}
The original datasets are from TCGA Broad GDAC Firehose (\url{https://gdac.broadinstitute.org/}). 
The harmonized datasets used in our study and the code for our model are available upon publication. 

\begin{acks}
This work is partially supported by the National Institute of General Medical Sciences of the National Institutes of Health (NIH/NIGMS) under Award P20GM104420. 
\end{acks}


\bibliographystyle{main}
\bibliography{main}

\end{document}